\documentclass[11pt]{article}
\usepackage{acl}

\usepackage{times}
\usepackage{CJKutf8}
\usepackage{latexsym}
\usepackage[T1]{fontenc}
\usepackage[utf8]{inputenc}
\usepackage{microtype}
\usepackage{inconsolata}
\usepackage{multirow}
\usepackage{amsmath,amsfonts,bm}

\def\eqref#1{equation~\ref{#1}}
\def\1{\bm{1}}

\DeclareMathAlphabet{\mathsfit}{\encodingdefault}{\sfdefault}{m}{sl}
\SetMathAlphabet{\mathsfit}{bold}{\encodingdefault}{\sfdefault}{bx}{n}

\usepackage{graphicx}
\usepackage{subcaption}
\usepackage{booktabs}
\usepackage{tabularx}
\usepackage{enumitem}
\usepackage{xcolor}
\usepackage{tikz}
\usetikzlibrary{arrows.meta,positioning,fit,shapes.misc}
\usepackage[ruled,vlined,linesnumbered]{algorithm2e}
\usepackage{newunicodechar}
\newunicodechar{≈}{\ensuremath{\approx}}
\usepackage{hyperref}
\usepackage{url}
\usepackage[table]{xcolor}
\usepackage{threeparttable}

\title{Redefining Machine Simultaneous Interpretation: From Incremental Translation to Human-Like Strategies}

\author{
  Qianen Zhang \\
  The Chinese University of Hong Kong,\\ Shenzhen \\
  Shenzhen, Guangdong, China \\
  \texttt{122090753@link.cuhk.edu.cn}
  \And
  Zeyu Yang \\
  The Chinese University of Hong Kong,\\ Shenzhen \\
  Shenzhen, Guangdong, China \\
  \texttt{zeyuyang1@link.cuhk.edu.cn}
  \AND
  Satoshi Nakamura \\
  The Chinese University of Hong Kong,\\ Shenzhen \\
  Shenzhen, Guangdong, China \\
  Nara Institute of Science\\ and Technology \\
  Nara, Japan \\
  \texttt{snakamura@cuhk.edu.cn}
}

\begin{document}
\begin{CJK}{UTF8}{gbsn}
\maketitle
\thispagestyle{plain}
\pagestyle{plain}
\begin{abstract}
Simultaneous Machine Translation (SiMT) requires high-quality translations under strict real-time constraints, which traditional policies with only READ/WRITE actions cannot fully address. We extend the action space of SiMT with four adaptive actions: \textbf{Sentence\_Cut}, \textbf{Drop}, \textbf{Partial\_Summarization} and \textbf{Pronominalization}, which enable real-time restructuring, omission, and simplification while preserving semantic fidelity. We adapt these actions in a large language model (LLM) framework and construct training references through action-aware prompting. To evaluate both quality and word-level monotonicity, we further develop a latency-aware TTS pipeline that maps textual outputs to speech with realistic timing. Experiments on the ACL60/60 English-Chinese, English-German and English-Japanese benchmarks show that our framework consistently improves semantic metrics and achieves lower delay compared to reference translations and salami-based baselines. Notably, combining \textbf{Drop} and \textbf{Sentence\_Cut} leads to consistent improvements in the balance between fluency and latency.
 These results demonstrate that enriching the action space of LLM-based SiMT provides a promising direction for bridging the gap between human and machine interpretation.
\end{abstract}

\section{Introduction}

Simultaneous speech translation requires real-time translation with high quality, which poses unique challenges compared to offline machine translation (MT) due to the incompleteness of information. While traditional MT systems generate fluent and accurate translations by relying on complete source sentences, such a paradigm is unsuitable for simultaneous machine translation (SiMT), where incremental processing and low latency are mandatory. The central bottleneck of SiMT therefore lies in maintaining an optimal balance between translation quality and latency. To address this challenge, the system must be capable of deciding both \textit{when} and \textit{how} to translate under partial input.

A prospective approach is to learn from professional human interpreters, who employ adaptive strategies such as rephrasing, summarization, and selective omission to cope with simultaneity constraints while preserving the core meaning of the source speech \citep{doi-etal-2021-large}. For instance, the widely adopted \textit{salami technique} breaks the sentences into minimal segments that contains enough information for translation. This segmentation effectively reduces long-distance word reordering caused by syntactic divergence across languages, while preserving semantic fidelity. By applying this technique to SiMT, the system benefits from the monotonicity and improved word-order alignment \citep{makinae-etal-2024-simul}.

Most existing work on SiMT builds on traditional machine translation systems, where translation is generated incrementally and the quality–latency trade-off is controlled by READ/WRITE policies. Such formulations restrict the space of translation behaviors and struggle to capture higher-level strategies used by human interpreters, such as sentence segmentation, selective omission, and partial reordering. This motivates the incorporation of summarization- and omission-oriented operations into SiMT modeling. In this work, we adopt large language models (LLMs) as the translation backbone and exploit their capacity for partial summarization, flexible reordering, and discourse-level rewriting to generate translations that more closely resemble human interpreter output under simultaneous constraints.

Another obstacle lies in training data. Most SiMT systems adopt offline translations as references. Although such translations are fluent and semantically faithful, they are unsuitable as references for SiMT because of different generation patterns. Training on them biases models toward waiting for complete input, thus increasing latency and contradicting the real-time requirement. As a result, it is crucial to generate high-quality reference interpretations that align with simultaneous interpretation patterns while preserving semantic fidelity.

In this study, we propose a LLM-based framework for simultaneous machine interpretation that extends the conventional READ/WRITE paradigm with four adaptive actions: \textbf{Sentence\_Cut}, \textbf{Partial\_Summarization}, \textbf{Drop}, and \textbf{Pronominalization}. By expanding the action space, our approach shifts the focus of SiMT from purely incremental token emission to the explicit modeling of human-like interpretation strategies, while remaining within the standard SiMT formulation.

Our framework is built on decoder-only large language models such as GPT-4o and Qwen3-8B, and is evaluated against strong baselines including salami-based segmentation, the LLM-based TransLLaMA system, and prompting-based methods such as few-shot prompting. Although our experiments are conducted in a text-to-text setting, we view the proposed method as modeling the decoder component of a full SiMT system. Speech input can be incorporated through an upstream encoder in future work. 
%In addition, we develop a latency-aware text-to-speech (TTS) pipeline based on word alignment and source timestamps, which enables realistic simulation of interpreter behavior and supports accurate evaluation of quality and latency-related monotonicity.

Our main contributions are as follows.\footnote{Due to double-blind review, we do not release artifacts during the review phase. Upon acceptance, we will release the code.}

\setlength{\leftmargini}{0pt}
\begin{itemize}
    \item We propose a LLM-based SiMT framework introducing four novel actions (\textbf{Sentence\_Cut}, \textbf{Partial\_Summarization}, \textbf{Drop} and \textbf{Pronominalization}) integrated into a sequential decision-making process to balance translation quality and latency.
    \item We adapt offline translations into SiMT-like references using these actions, and produce training data that better reflects real-time interpretation constraints while preserving semantic fidelity.
    \item We systematically compare multiple LLM-based inference paradigms for SiMT to analyze their respective strengths in balancing translation quality and latency.
    % \item We developed a latency-aware TTS pipeline based on word alignment and source timestamps, enabling realistic simulation of interpreter behavior and synchronous evaluation of quality and latency.

\end{itemize}

\section{Related work}

Current SiMT systems are typically evaluated in terms of translation quality and latency. Translation quality is commonly measured using surface-based metrics such as BLEU \citep{papineniBLEUMethodAutomatic2001}, chrF \citep{popovic-2015-chrf}, and TER \citep{snover-etal-2006-study}, which rely on n-gram overlap and may undervalue semantically valid but lexically divergent translations. In contrast, neural metrics such as COMET \citep{DBLP:journals/corr/abs-2009-09025, rei-etal-2022-cometkiwi} compare meaning using pretrained multilingual encoders and correlate better with human judgments \citep{glushkova-etal-2023-bleu}. 

Latency is commonly evaluated using AL \citep{ma-etal-2019-stacl}, AP \citep{DBLP:journals/corr/ChoE16}, DAL \citep{DBLP:journals/corr/abs-1808-09943}, and LAAL \citep{papi2022over}, with more recent work proposing ATD to explicitly account for delays caused by long target segments \citep{kano2023averagetokendelaylatency}. Since BLEU depends on offline references and may encourage excessive waiting in SiMT, several recent studies on simultaneous and simulMT evaluation modify offline references or adopt salami-style segmentation to reduce long-distance reordering effects and better reflect real-time interpretation behavior \citep{DBLP:journals/corr/abs-2010-11247, makinae-etal-2024-simul, DBLP:journals/corr/abs-2110-05213}.

In terms of architecture, most SiMT systems adopt encoder–decoder frameworks, where a policy determines when to READ source tokens and WRITE target tokens to balance translation quality and latency. Early work explored fixed policies such as wait-k \citep{ma-etal-2019-stacl} and segmentation-based approaches \citep{odaOptimizingSegmentationStrategies2014}, while later studies proposed adaptive policies that adjust READ/WRITE decisions based on source content or model predictions \citep{DBLP:journals/corr/abs-1906-05218, oda-etal-2015-syntax}. Notably, \citet{oda-etal-2015-syntax} introduced one of the earliest syntax-informed approaches by predicting unseen constituents. More recent simulMT research has explored decoder-only architectures and large pretrained language models for incremental and low-latency translation \citep{guo2024sillm, koshkin2024transllamallmbasedsimultaneoustranslation, wang-etal-2024-simultaneous}.

Distinct from previous work, we propose a SiMT framework based on LLMs empowered by four novel actions (\textbf{Sentence\_Cut}, \textbf{Partial\_Summarization}, \textbf{Drop} and \textbf{Pronominalization}) to balance quality and latency by emulating the adaptive strategies of professional interpreters, which we describe below.

\section{Method}
\begin{figure}[t]
    \centering
    \includegraphics[width=0.85\linewidth]{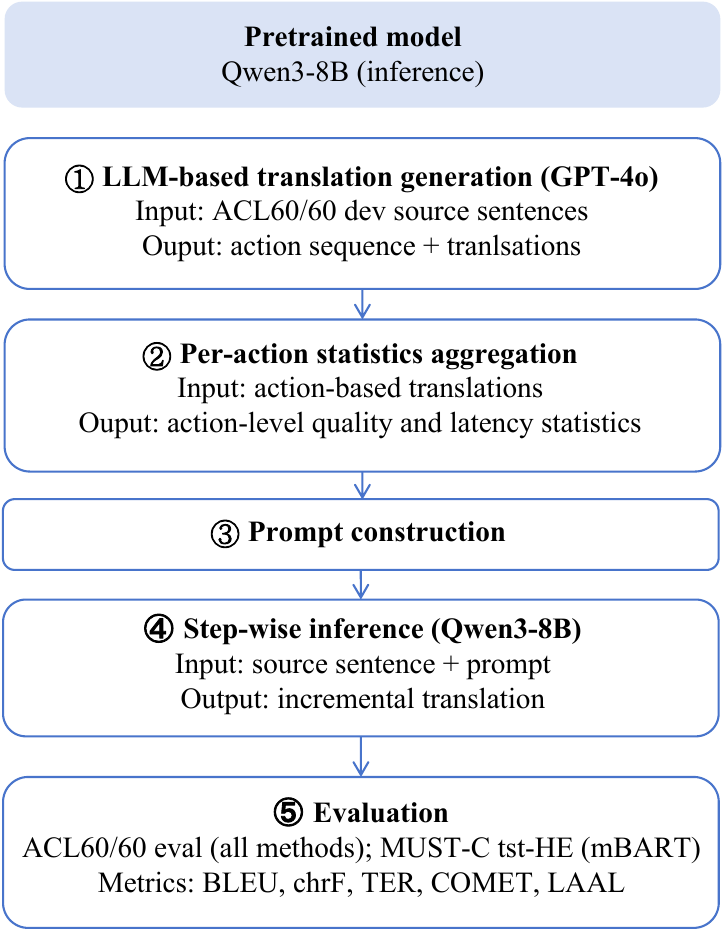}
    \caption{Block diagram of the inference procedure. We first obtain per-action translation BLEU and LAAL statistics by evaluating LLM generated translations of ACL60/60 dev set. These statistics are provided for the LLM, which is prompted to translate the input sentence by choosing an action at each step according to the statistics.}
    \label{fig:method}
\end{figure}

\begin{figure*}
    \centering
    \includegraphics[width=0.9\linewidth]{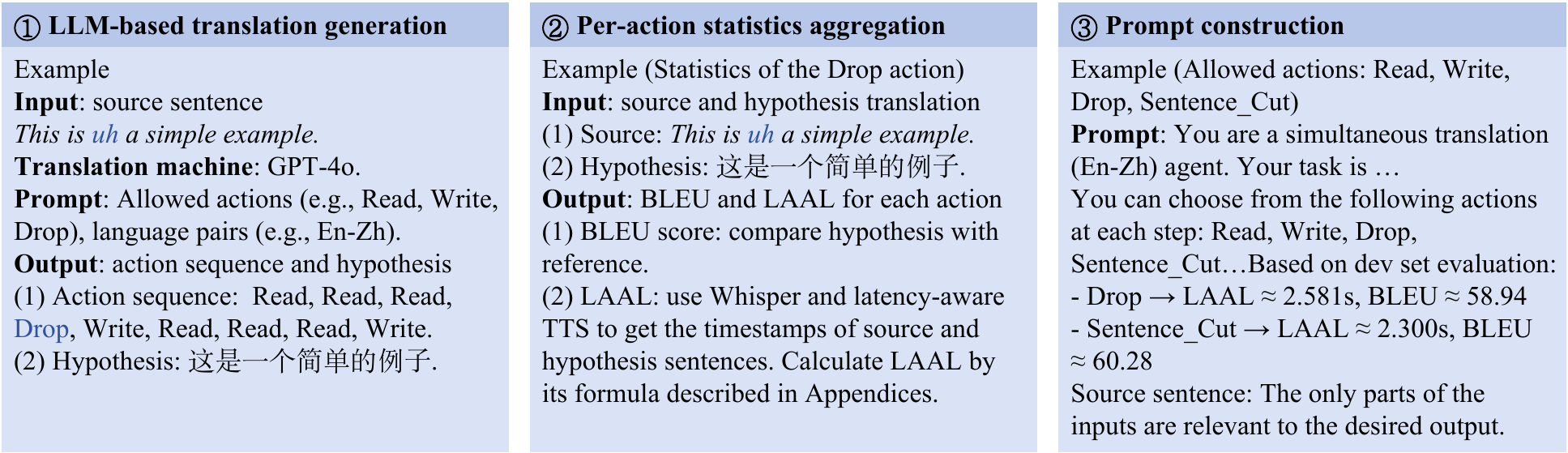}
    \caption{Examples accompanying the inference pipeline. The figure provides simplified input–output examples for LLM-based action generation, explains how translation quality (BLEU) and latency (LAAL) are obtained, and shows a sample inference prompt that combines allowed actions with action-level statistics. This figure is intended for clarification and does not reflect full inference traces.}
    \label{fig:example}
\end{figure*}

We propose a LLM-based framework for SiMT that formulates generation as a step-wise, action-guided process. The method consists of three main components: (1) an extended action space that abstracts human interpreter strategies beyond the conventional READ/WRITE paradigm, (2) a prompt-based inference procedure that enables a decoder-only LLM to select actions dynamically during generation, and (3) a latency-aware TTS pipeline for evaluating word-level monotonicity in the time domain. Figure~\ref{fig:method} provides an overview of the inference procedure, and Figure~\ref{fig:example} illustrates the prompt design and step-wise generation process.

\subsection{Extended action space} 
Conventional SiMT policies are limited to two actions: READ and WRITE . Although the optimization of policies can improve quality-latency trade-off by making better decisions of when to commit WRITE actions, they cannot fully capture the techniques developed by human interpreters, which make translations more fluent and accurate.

Building on this idea, we generalize such human interpreter techniques into four new actions that can be dynamically invoked by LLMs during generation, enabling real-time application beyond static reference adaptation.

% \begin{figure*}[t]
%     \centering
%     \includegraphics[width=0.75\linewidth]{fig/actions_illustration2.pdf}
%     \caption{New actions and examples in En-Zh SiMT. Each action is illustrated with an English source sentence, a literal reference translation, and an adapted version with the corresponding action applied. Highlighted spans indicate the parts of the sentence that are treated differently in the adapted translation compared to the reference (e.g., splitting, omission, summarization). This visualization shows how the original English segment is restructured, modified, or condensed in Chinese translation.}
%     \label{fig:New actions}
% \end{figure*}

\setlength{\leftmargini}{0pt}
\begin{itemize}
    \item \textbf{Sentence\_Cut} Split long or syntactically complex clauses into shorter, grammatical sentence segments. For example, the sentence in ACL60/60 dataset \textit{``At the position of at the pool party with Barack Obama, we got a graph with the right nodes on the person and the event subject, but guess the wrong timing information''} can be split before ``with'' since it connects two semantically complete parts. By inserting appropriate punctuation and connective words (which LLMs are able to supply), the sentence is divided into two fluent units, reducing reordering and improving latency.

    \item \textbf{Drop} Remove only truly non-informative content (e.g., ``uh'', ``you know''), repeated words, or self-corrections. For instance, in \textit{``These are the morphology level, these are the morphology level embeddings,''} the phrase \textit{``these are the morphology level''} is repeated without adding new meaning. Applying \textbf{Drop} removes the redundancy and yields a cleaner, semantically accurate translation.

    \item \textbf{Partial\_Summarization} Combine or simplify semantically equivalent or repetitive expressions while preserving the original meaning and tone (e.g., speculation, politeness). This is useful when multiple clauses convey essentially the same information. For example, in \textit{“And here you have the number of spans that were labeled as English and the spans that were labeled as other borrowings and how many of them were unique,”} both clauses share the subject \textit{“the number of spans.”} Summarization condenses the sentence into a more concise form, improving readability and reducing latency without loss of meaning.

    \item \textbf{Pronominalization} Replace repeated or already mentioned noun phrases with pronouns only if referents are unambiguous. In two consecutive sentences \textit{``Lexical borrowing is a type of linguistic borrowing um which is basically reproducing in one language patterns of other languages. There are however some differences between lexical borrowing and code-switching''}, the phrase \textit{``lexical borrowing''} is repeated. Since no ambiguity would be caused if we replace the second phrase into a pronoun like ``it'', and the two phrases are close enough to each other, we can apply \textbf{Pronominalization} to convert this long phrase into a pronoun for more fluent expression.

\end{itemize}

These actions are applied at each decision point, and enables the model to adjust its behavior according to available context and the latency constraints. Training references are prepared by prompting GPT-4o to generate translations under different action combinations. Examples in En-Zh and En-De are shown in Appendix~\ref{app: new action}, and detailed experiment settings can be found in \ref{gpt4o prompt}.

\subsection{Prompt-based action-guided inference}
\label{Inference}

Figure~\ref{fig:method} illustrates the overall inference pipeline of the proposed framework. We formulate simultaneous interpretation as a step-wise decision process, where an LLM incrementally generates translations by explicitly selecting actions from the action space.

The inference procedure consists of three stages. First, we perform per-action statistics aggregation on the ACL60/60 development set. Using GPT-4o, we generate translations by applying each individual action as well as selected action combinations (sub-figure ① in Figure~\ref{fig:example}). For each action, we compute translation quality (BLEU) and word-level monotonicity measured by LAAL. These statistics serve as empirical indicators of the quality-latency trade-offs associated with different actions, as indicated in sub-figure ② in Figure~\ref{fig:example}.

Second, we construct an action-aware inference prompt, as illustrated in sub-figure ③ in Figure~\ref{fig:example}. The prompt specifies the allowed action set, describes the interpretation task, and incorporates the aggregated BLEU and LAAL statistics for each action. By exposing the model to these statistics, the prompt explicitly informs the LLM of the expected consequences of choosing different actions during generation.

Finally, we perform step-wise action selection at inference time. Given the source word sequence and the constructed prompt, the LLM selects an action at each decision point and generates the corresponding translation segment. This process allows the model to dynamically balance translation quality and latency based on both the input context and the action-level statistics.

\subsection{TTS-based LAAL via a Latency-aware TTS Pipeline}

We evaluate word-level monotonicity using a time-based variant of Length-Adaptive Average Lagging, referred to as \emph{TTS-based LAAL}, which measures target emission behavior on a real-time axis and can be directly interpreted in the context of simultaneous interpretation. 

In speech-based systems, such metrics can be computed directly from audio timestamps; however, target-side speech timing is unavailable in text-only setting. To address this, we introduce a latency-aware TTS pipeline that maps translations to speech with causally consistent timing, enabling the computation of TTS-based LAAL.  Implementation details of the latency-aware TTS pipeline are provided in Appendix~\ref{app: TTS}.

\section{Experiment and analysis}

\subsection{Data}
All experiments are conducted in a text-only simultaneous translation setting, where the input is an English word-level transcript and the output is incrementally generated target-side text.

We use two benchmarks with clearly separated roles: MUST-C \citep{di-gangi-etal-2019-must} and ACL60/60 \citep{salesky-etal-2023-evaluating}. The MUST-C corpus is used for supervised encoder-decoder experiments under different reference styles. For En-Zh and En-De, we sample 25k instances from the MUST-C training set and construct three target-side variants with identical sources: the original references, salami-based references, and action-adapted references produced with our prompting strategies. Model validation is performed on the MUST-C \texttt{tst-HE} set and ACL60/60 evaluation (eval) set.

All LLM-based experiments are conducted on the ACL60/60 benchmark. The ACL60/60 development (dev) set is used to collect per-action translation quality and word-level monotonicity statistics and to select demonstration examples for prompting-based methods. The ACL60/60 evaluation (eval) set is used for all final evaluations of LLM-based approaches. Inference experiments cover three language pairs: En-Zh, En-De, and En-Ja.

\subsection{Backbone models}
We consider two types of backbone models in our experiments. For supervised encoder-decoder training, we use mBART50 (\texttt{facebook/mbart-large-50-many-to-many-mmt}) \citep{liu-etal-2020-multilingual-denoising}. For all LLM-based methods, including TransLLaMA-style fine-tuning, few-shot prompting, and step-wise action-based inference, we use Qwen3-8B as the base decoder-only language model. In addition, GPT-4o is used during development to generate action-based translations for collecting per-action quality and monotonicity statistics, but it is not used for final evaluation.

\subsection{Evaluation metrics} Translation quality was measured by BLEU, chrF, TER, and neural-based metrics COMET-da and COMET-KIWI. Latency was evaluated using LAAL (Length-Adaptive Average Lagging).

\subsection{Baseline systems}
To evaluate the effectiveness of the proposed action-based framework, we compare against several representative baselines that cover supervised encoder--decoder training, LLM-based supervised adaptation, and prompting-based inference. All baseline systems are evaluated under the same text-only simultaneous translation setting and using identical evaluation metrics. Detailed experimental settings for all baselines are provided in Appendix~\ref{app: experimental setup}.

\subsubsection{mBART50 fine-tuning}
As a supervised encoder-decoder baseline, we fine-tune mBART50 on parallel data derived from the MUST-C training set. Separate models are fine-tuned under three reference styles: (1) the original MUST-C reference translations, (2) salami-based translations generated using salami prompting, and (3) action-based translations generated with our full action prompt. All mBART50 models share identical source inputs and training configurations, allowing us to isolate the effect of reference adaptation. These models are evaluated on the MUST-C \texttt{tst-HE} set and the ACL60/60 evaluation set.

\subsubsection{TransLLaMA}
We include TransLLaMA as a strong LLM-based baseline that adapts decoder-only language models to simultaneous translation through supervised fine-tuning. Following \citet{koshkin2024transllamallmbasedsimultaneoustranslation}, training data are causally aligned by inserting \texttt{<WAIT>} tokens into the target sequences, enabling the model to implicitly learn when to generate or wait without relying on an explicit action-selection mechanism. TransLLaMA is fine-tuned on the ACL60/60 development set under three reference styles: (1) ACL60/60 dev set reference translations, (2) salami-based translations, and (3) action-based translations. Results are evaluated on the ACL60/60 evaluation set.

\subsubsection{Few-shot prompting}
Few-shot prompting is used as a parameter-free baseline that leverages the in-context learning capability of large language models. We construct prompts by providing three demonstration examples selected from the ACL60/60 development set, covering reference, salami-based, and action-adapted translations. These demonstrations guide the model toward different simultaneous translation behaviors without updating model parameters. Few-shot prompting is performed using the same backbone model as other LLM-based methods and evaluated on the ACL60/60 evaluation set. Detailed prompt construction is shown in Appendix \ref{app:few-shot prompt}.

\subsection{Main results} 
% \begin{table*}[t]
% \centering
% \small
% \caption{Model adaptation results under three supervision settings on ACL60/60 eval set (En-Zh).
% We report BLEU, COMET-KIWI, and latency (LAAL, seconds$\downarrow$).}
% \label{tab:simplified_table1}

% \begin{threeparttable}
% \begin{tabular}{lccc|ccc|ccc}
% \toprule
% & \multicolumn{3}{c|}{\textbf{Salami-based}} 
% & \multicolumn{3}{c|}{\textbf{Action-adapted}} 
% & \multicolumn{3}{c}{\textbf{ACL60/60 reference}} \\
% \cmidrule(lr){2-4} \cmidrule(lr){5-7} \cmidrule(lr){8-10}

% \textbf{Method} 
% & BLEU & COMET & LAAL$\downarrow$ 
% & BLEU & COMET & LAAL$\downarrow$ 
% & BLEU & COMET & LAAL$\downarrow$ \\
% \midrule
% TransLLaMA 
% & \textbf{57.66} & 0.7950 & \textbf{2.376}
% & \textbf{58.50} & 0.8053 & \textbf{2.360}
% & \textbf{57.66} & 0.8000 & \textbf{2.934} \\

% Few-shot prompting
% & 55.49 & \textbf{0.7984} & 2.456
% & 55.80 & \textbf{0.8080} & 2.395
% & 55.79 & \textbf{0.8079} & 2.945 \\
% \bottomrule
% \end{tabular}

% \begin{tablenotes}[flushleft]
% \footnotesize
% \item COMET refers to COMET-KIWI. Lower LAAL indicates lower latency.
% \end{tablenotes}
% \end{threeparttable}
% \end{table*}

\begin{table*}[t]
\small
\centering
\caption{mBART50 supervised adaptation under different training data and evaluation benchmarks. 
\emph{Reference} denotes the original offline MUST-C translations; \emph{salami-based} and \emph{action-based} refer to targets adapted for simultaneous translation using salami segmentation and our proposed action-based prompting, respectively. 
LAAL (s) denotes TTS-based LAAL measured on a real-time axis using the latency-aware TTS pipeline, while LAAL (tok) is the conventional token-level LAAL evaluated by SimulEval \citep{ma-etal-2020-simuleval}.}

\label{tab:mbart}

% ======================================================
% En-Zh
% ======================================================
\begin{subtable}{\linewidth}
\centering
\caption{English-Chinese (En-Zh): Evaluation on MUST-C \texttt{tst-HE}}
\begin{threeparttable}
\begin{tabular}{lccccccc}
\toprule
\textbf{Training data} &
\textbf{BLEU} &
\textbf{chrF} &
\textbf{TER$\downarrow$} &
\textbf{COMET-da} &
\textbf{COMET-KIWI} &
\textbf{LAAL (s)$\downarrow$} &
\textbf{LAAL (tok)$\downarrow$} \\
\midrule
MUST-C (reference)            
& 21.29 & 14.34 & 101.80 & 0.7525 & 0.7292 & 2.650 &  4.572\\

MUST-C + salami-based         
& 23.03 & 15.35 & 109.48 & 0.7903 & 0.7653 & 2.656 & 4.573 \\

MUST-C + action-based         
& \textbf{24.34} & \textbf{16.39} & \textbf{100.17} & \textbf{0.8002} & \textbf{0.7900} & \textbf{2.438} & \textbf{4.544} \\
\bottomrule
\end{tabular}
\end{threeparttable}
\end{subtable}

\vspace{0.6em}

\begin{subtable}{\linewidth}
\centering
\caption{English-Chinese (En-Zh): Evaluation on ACL60/60 eval}
\begin{threeparttable}
\begin{tabular}{lccccccc}
\toprule
\textbf{Training data} &
\textbf{BLEU} &
\textbf{chrF} &
\textbf{TER$\downarrow$} &
\textbf{COMET-da} &
\textbf{COMET-KIWI} &
\textbf{LAAL (s)$\downarrow$} &
\textbf{LAAL (tok)$\downarrow$} \\
\midrule
MUST-C (reference)            
& 40.57 & 26.29 & 137.77 & 0.8092 & 0.7377 & 2.587 & 4.523 \\

MUST-C + salami-based         
& 38.80 & 26.84 & 169.21 & 0.8260 & 0.7564 & 2.568 & 4.501 \\

MUST-C + action-based         
& \textbf{45.07} & \textbf{29.97} & \textbf{113.54} & \textbf{0.8436} & \textbf{0.7801} & \textbf{2.533} & \textbf{4.443} \\
\bottomrule
\end{tabular}
\end{threeparttable}
\end{subtable}

\vspace{0.8em}

% ======================================================
% En-De
% ======================================================
\begin{subtable}{\linewidth}
\centering
\caption{English-German (En-De): Evaluation on MUST-C \texttt{tst-HE}}
\begin{threeparttable}
\begin{tabular}{lccccccc}
\toprule
\textbf{Training data} &
\textbf{BLEU} &
\textbf{chrF} &
\textbf{TER$\downarrow$} &
\textbf{COMET-da} &
\textbf{COMET-KIWI} &
\textbf{LAAL (s)$\downarrow$} &
\textbf{LAAL (tok)$\downarrow$} \\
\midrule
MUST-C (reference)            
& 24.90 & 51.17 & 65.45 & 0.7834 & 0.7879 & 1.251 & 4.061 \\

MUST-C + salami-based         
& 25.47 & 51.72 & 66.12 & 0.7865 & 0.7943 & 1.223 & 3.681 \\

MUST-C + action-based         
& \textbf{25.69} & \textbf{51.84} & \textbf{65.20} & \textbf{0.7887} & \textbf{0.7944} & \textbf{1.215} & \textbf{3.648} \\
\bottomrule
\end{tabular}
\end{threeparttable}
\end{subtable}

\vspace{0.6em}

\begin{subtable}{\linewidth}
\centering
\caption{English-German (En-De): Evaluation on ACL60/60 eval}
\begin{threeparttable}
\begin{tabular}{lccccccc}
\toprule
\textbf{Training data} &
\textbf{BLEU} &
\textbf{chrF} &
\textbf{TER$\downarrow$} &
\textbf{COMET-da} &
\textbf{COMET-KIWI} &
\textbf{LAAL (s)$\downarrow$} &
\textbf{LAAL (tok)$\downarrow$} \\
\midrule
MUST-C (reference)            
& 30.57 & 56.90 & 54.70 & \textbf{0.7890} & 0.7646 & 1.210 & 4.006 \\

MUST-C + salami-based         
& 35.58 & 60.21 & 50.81 & 0.7884 & 0.7669 & 1.191 & 3.681 \\

MUST-C + action-based         
& \textbf{36.02} & \textbf{60.74} & \textbf{50.09} & \textbf{0.7890} & \textbf{0.7687} & \textbf{1.189} & \textbf{3.605} \\
\bottomrule
\end{tabular}
\end{threeparttable}
\end{subtable}

\end{table*}

\begin{table*}
\centering
\small
\caption{
Overview of evaluated systems and inference results on the ACL60/60 eval set.
\textbf{(a)} Definition of systems evaluated in this work; each system differs in model backbone, tuning strategy, supervision data, and inference-time prompting.
\textbf{(b)} English-Chinese (En-Zh),
\textbf{(c)} English-German (En-De),
and \textbf{(d)} English-Japanese (En-Ja) inference results, reported for systems ①-⑩ defined in (a).
}
\label{tab:system_overview}

% =========================
% (a) System definition
% =========================
\begin{subtable}{\textwidth}
\centering
\caption{System definition}
\begin{tabular}{clllll}
\toprule
\textbf{ID} & \textbf{System} & \textbf{Tuning method} & \textbf{Tuning data} & \textbf{Inference prompt} \\
\midrule
① & TransLLaMA (Salami)    & Fine-tuning & Dev set (Salami refs)        & -- \\
② & TransLLaMA (Reference) & Fine-tuning & Dev set (Reference)          & -- \\
③ & TransLLaMA (Action)    & Fine-tuning & Dev set (Action-adapted refs)& -- \\
④ & Few-shot (Salami)      & Prompt tuning & Dev set (Salami refs)      & Few-shot demonstrations \\
⑤ & Few-shot (Reference)   & Prompt tuning & Dev set (Reference)        & Few-shot demonstrations \\
⑥ & Few-shot (Action)      & Prompt tuning & Dev set (Action-adapted refs)& Few-shot demonstrations \\
⑦ & Salami                 & Fine-tuning & Dev set (Salami refs)        & Salami prompt \\
⑧ & Step-by-step (Actions) & -- & -- & Incremental action selection \\
⑨ & Step-by-step (Drop / Cut) & -- & -- & Incremental drop / cut selection \\
⑩ & Reference              & -- & -- & -- \\
\bottomrule
\end{tabular}
\end{subtable}

\vspace{0.4em}

% =========================
% (b) En-Zh
% =========================
\begin{subtable}{\textwidth}
\centering
\caption{English-Chinese (En-Zh)}
\begin{tabular}{p{2.8cm}p{0.8cm}cccccc}
\toprule
& \textbf{ID} & \textbf{BLEU} & \textbf{chrF} & \textbf{TER$\downarrow$} & \textbf{COMET-da} & \textbf{COMET-KIWI} & \textbf{LAAL (s)$\downarrow$} \\
\midrule
\multirow{3}{*}{\textbf{TransLLaMA}}
& ① & 57.66 & 41.36 & 97.16 & 0.8768 & 0.7950 & 2.376 \\
& ② & 57.66 & 41.27 & 97.16 & 0.8852 & 0.8000 & 2.934 \\
& ③ & 58.50 & 41.61 & \textbf{96.72} & 0.8826 & 0.8053 & 2.360 \\
\multirow{3}{*}{\textbf{Few-shot}}
& ④ & 55.49 & 50.11 & 106.11 & 0.8779 & 0.7984 & 2.456 \\
& ⑤ & 55.79 & 39.11 & 110.77 & 0.8856 & 0.8040 & 2.945 \\
& ⑥ & 55.80 & \textbf{50.31} & 105.90 & 0.8843 & 0.8079 & 2.395 \\
\midrule
\multicolumn{2}{l}{⑦ Salami} & 57.21 & 40.57 & 110.04 & 0.8705 & 0.7846 & 2.521 \\
\multicolumn{2}{l}{⑧ Step-wise (Actions)} & 62.44 & 46.80 & 126.20 & \textbf{0.9002} & 0.8020 & 2.281 \\
\multicolumn{2}{l}{⑨ Step-wise (Drop / Cut)} & \textbf{62.84} & 44.06 & 104.80 & 0.8891 &\textbf{ 0.8080} & \textbf{2.118} \\
\rowcolor{black!5}
\multicolumn{2}{l}{⑩ Reference} & 100.00 & 100.00 & 0.00 & 0.9582 & 0.8029 & 2.848 \\
\bottomrule
\end{tabular}
\end{subtable}

\vspace{0.35em}

% =========================
% (c) En-De
% =========================
\begin{subtable}{\textwidth}
\centering
\caption{English-German (En-De)}
\begin{tabular}{p{2.8cm}p{0.8cm}cccccc}
\toprule
& \textbf{ID} & \textbf{BLEU} & \textbf{chrF} & \textbf{TER$\downarrow$} & \textbf{COMET-da} & \textbf{COMET-KIWI} & \textbf{LAAL (s)$\downarrow$} \\
\midrule
\multirow{3}{*}{\textbf{TransLLaMA}}
& ① & 39.12 & 59.67 & 46.68 & 0.8125 & 0.7798 & 1.187 \\
& ② & 39.52 & 59.61 & 47.09 & 0.8198 & 0.7839 & 1.193 \\
& ③ & 40.31 & 60.43 & 45.87 & 0.8235 & 0.7842 & \textbf{1.173} \\
\multirow{3}{*}{\textbf{Few-shot}}
& ④ & 38.62 & 62.74 & 47.83 & 0.8245 & 0.7859 & 1.192 \\
& ⑤ & 38.28 & 62.33 & 48.00 & 0.8301 & 0.7876 & 1.196 \\
& ⑥ & 38.92 & 63.17 & 47.85 & 0.8241 & 0.7856 & 1.180 \\
\midrule
\multicolumn{2}{l}{⑦ Salami} & 47.48 & 69.87 & 38.94 & 0.8534 & 0.8102 & 1.182 \\
\multicolumn{2}{l}{⑧ Step-wise (Actions)} & 47.80 & 70.08 & 38.72 & 0.8541 & 0.8108 & \textbf{1.173} \\
\multicolumn{2}{l}{⑨ Step-wise (Drop / Cut)} & \textbf{49.97} & \textbf{70.96} & \textbf{37.38} & \textbf{0.8594} & \textbf{0.8111} & 1.180 \\
\rowcolor{black!5}
\multicolumn{2}{l}{⑩ Reference} & 100.00 & 100.00 & 0.00 & 0.9511 & 0.8048 & 2.507 \\
\bottomrule
\end{tabular}
\end{subtable}

\vspace{0.35em}

% =========================
% (d) En-Ja
% =========================
\begin{subtable}{\textwidth}
\centering
\caption{English-Japanese (En-Ja)}
\begin{tabular}{p{2.8cm}p{0.8cm}cccccc}
\toprule
& \textbf{ID} & \textbf{BLEU} & \textbf{chrF} & \textbf{TER$\downarrow$} & \textbf{COMET-da} & \textbf{COMET-KIWI} & \textbf{LAAL (s)$\downarrow$} \\
\midrule
\multirow{3}{*}{\textbf{TransLLaMA}}
& ① & 53.12 & 30.11 & 98.35 & 0.8735 & 0.8313 & 2.938 \\
& ② & 54.88 & 30.48 & 98.58 & 0.8789 & 0.8429 & 3.085 \\
& ③ & 54.08 & 30.69 & \textbf{97.72} & 0.8837 & 0.8351 & 2.927 \\
\multirow{3}{*}{\textbf{Few-shot}}
& ④ & 52.27 & 34.93 & 108.25 & 0.8926 & 0.8379 & 3.049 \\
& ⑤ & 52.91 & 35.02 & 104.72 & 0.8934 & 0.8423 & 3.146 \\
& ⑥ & 52.78 & \textbf{35.08} & 107.31 & 0.8913 & 0.8428 & 2.915 \\
\midrule
\multicolumn{2}{l}{⑦ Salami} & 50.16 & 31.98 & 106.37 & 0.8854 & 0.8295 & 2.955 \\
\multicolumn{2}{l}{⑧ Step-wise (Actions)} & 55.31 & 34.33 & 101.42 & \textbf{0.8998} & 0.8439 & \textbf{2.524} \\
\multicolumn{2}{l}{⑨ Step-wise (Drop / Cut)} & \textbf{55.33} & 34.49 & 99.06 & 0.8913 & 0.\textbf{8447} & 2.613 \\
\rowcolor{black!5}
\multicolumn{2}{l}{⑩ Reference} & 100.00 & 100.00 & 0.00 & 0.9721 & 0.8447 & 3.298 \\
\bottomrule
\end{tabular}
\end{subtable}

\end{table*}
\subsubsection{mBART50 fine-tuning} Table~\ref{tab:mbart} reports the performance of mBART50 under different supervision styles on the English-Chinese (En-Zh) and English-German (En-De) language pairs, evaluated on both MUST-C \texttt{tst-HE} and the ACL60/60 eval set. Training on action-based targets consistently outperforms the original MUST-C references and salami-based translations across most metrics, indicating that reference adaptation is beneficial for supervised encoder-decoder models in simultaneous translation settings. In particular, action-based supervision yields the strongest overall performance on both benchmarks, achieving higher BLEU and COMET scores while also reducing latency as measured by LAAL. This suggests that action-style targets encourage more monotonic and semantically faithful generation without sacrificing fluency.

\begin{table*} \centering \small \caption{System-level ranking induced by different word monotonicity proxies, computed from mBART50 inference results. Lower is better for LAAL-based metrics, while higher is better for Spearman correlation. The table reports relative system ordering rather than absolute scores.} \label{tab:ranking_monotonicity} \begin{tabular}{lcc} \toprule \textbf{Metric} & \textbf{En--Zh} & \textbf{En--De} \\ \midrule TTS-based LAAL (s) & Action $<$ Salami $<$ MUST-C & Action $<$ Salami $<$ MUST-C \\ Token-level LAAL & Action $<$ Salami $<$ MUST-C & Action $<$ Salami $<$ MUST-C \\ Spearman correlation & Salami $<$ Action $<$ MUST-C & Salami $<$ Action $<$ MUST-C \\ \bottomrule \end{tabular} \end{table*}

\subsubsection{LLM-based model experiments} We report inference results on the ACL60/60 eval set for three language pairs (En-Zh, En-De, and En-Ja), comparing TransLLaMA-style supervised fine-tuning, few-shot prompting, and step-wise action-based inference under different reference settings in Table \ref{tab:system_overview}. The results show that the choice of supervision or prompting style (reference, salami-based, or action-based) has a consistent impact across methods. For both TransLLaMA and few-shot prompting, action-based references generally lead to higher semantic scores and lower latency compared to reference and salami-based settings, indicating that action-aware supervision better aligns model behavior with the requirements of simultaneous translation.

When comparing different inference paradigms, step-wise action-based inference consistently achieves a better overall quality–latency trade-off than TransLLaMA and few-shot prompting. In particular, the step-wise variants obtain higher BLEU and COMET scores while maintaining lower or comparable latency across all three language pairs. This suggests that explicitly selecting actions during inference provides finer-grained control over incremental generation than relying solely on supervised fine-tuning or static prompting. We further verify that the LLM adapts its action choices to the provided per-action statistics; see Appendix~\ref{app: adaptive behavior}.
While TransLLaMA remains competitive in latency-oriented configurations and few-shot prompting achieves strong semantic performance in some cases, both methods are generally outperformed by step-wise inference when quality and latency are considered jointly.

Across three language pairs, inference guided by action choices consistently outperforms salami segmentation, TransLLaMA fine-tuning and few-shot prompting in most quality metrics, and even yields COMET-KIWI scores comparable to the reference translations. Notably, the action combination ``\textbf{Drop + Sentence\_Cut}'', which achieved the lowest latency on the dev sets, also leads to the best trade-off on the eval sets. For En-Zh and En-Ja, it delivers the highest BLEU and COMET-KIWI while substantially reducing LAAL. For En-De, it improves all quality scores simultaneously while maintaining competitive latency.

These results suggest a shared trend across three language pairs: action-aware inference is better at balancing semantic fidelity and latency than other inference methods. The consistent advantage of ``\textbf{Drop + Sentence\_Cut}'' can be explained by their complementary effects: \textbf{Drop} removes redundant or filler material, reducing delay, while \textbf{Sentence\_Cut} alleviates long-distance reordering by segmenting complex clauses. Together, they enable translations that are both more fluent and faster, aligning with human interpreter strategies. 

For example, given the source sentence \textit{``In other words, it cannot be used for many projects on GitHub''}, the translation obtained with salami segmentation was ``换句话说,它不能被使用在许多项目中在GitHub上'' (BLEU 49.25), where ``GitHub上'' was placed at the end, resulting in an expression that does not conform to natural Chinese usage and providing little benefit in latency reduction. In contrast, inference with \textbf{Sentence\_Cut+Drop} generated ``换句话说,它不能用于GitHub上的许多项目'' (BLEU 100.00), which is both fluent and faithful. 
% As another case, for the sentence ``For those samples without unused quantities, so the overall performance is actually higher than the, the performance is actually higher than the overall performance'', the salami-based output preserved the redundant phrase (BLEU 52.61), while the \textbf{Drop} action effectively removed it, yielding ``对于那些没有未使用数量的样本,所以整体性能实际上高于整体性能'' (BLEU 66.89). This not only improved translation quality but also reduced latency since the output was shorter.

\subsection{Analysis of TTS-based LAAL as a Proxy for Word Monotonicity}
\label{sec:laal_ranking}

To assess whether TTS-based LAAL provides a stable and interpretable signal of word-level monotonicity, we analyze its system-level behavior and compare the induced rankings with those from existing monotonicity and latency proxies.

We compare system rankings produced by TTS-based LAAL, token-level LAAL, and alignment-based Spearman correlation, which has been used as a proxy for word monotonicity in prior work \citep{makinae-etal-2024-simul}. Table~\ref{tab:ranking_monotonicity} summarizes the results across language pairs.

Across both En-Zh and En-De, TTS-based LAAL yields identical system rankings to token-level LAAL, indicating that projecting decoding-based latency onto a real-time axis preserves relative monotonicity while expressing delay in seconds. In contrast, Spearman correlation sometimes produces different rankings, suggesting that it captures complementary word-order effects beyond emission timing. Additional analysis relating TTS-based LAAL to end-to-end delay metrics is provided in Appendix~\ref{app: laal_analysis}.

\section{Conclusions}

We presented a LLM-based framework for simultaneous machine interpretation that extends the traditional READ/WRITE paradigm with a set of interpreter-inspired actions. By enabling new actions \textbf{Sentence\_Cut}, \textbf{Partial\_Summarization}, \textbf{Drop}, and \textbf{Pronominalization}, the framework allows large language models to generate translations that better reflect human interpreting strategies under simultaneous constraints. We also introduced a latency-aware TTS pipeline that supports time-based evaluation of word-level monotonicity using TTS-based LAAL.

Experiments across supervised encoder-decoder training and LLM-based inference show that action-aware supervision and inference consistently lead to better trade-offs between translation quality and monotonicity than reference-based or salami-style approaches, with step-wise action selection yielding the strongest overall performance. More broadly, our findings suggest that simultaneous machine interpretation can be formulated as action-driven, human-like decision making at decoding time, offering a flexible alternative to policy-based incremental translation.

\section{Limitations and Future Work}
This work studies simultaneous machine interpretation from a text-to-text perspective, with a focus on action-based generation strategies. While speech input is not explicitly modeled, the proposed framework is intended to operate as the decoding component of a full SiMT system, where speech signals can be incorporated through an upstream encoder. Extending the framework to an end-to-end SiMT setting that jointly models speech input and action-based decoding is left for future work.

\bibliographystyle{acl_natbib}
\bibliography{reference}

\appendix
\section{Appendix}
\label{sec:appendix}

\subsection{New action examples}
\label{app: new action}
We provide concrete English-German (En-De) and English-Chinese (En-Zh) examples to demonstrate the functions of each new action. The source sentences and reference translations are selected from the ACL60/60 development set, and the action-based translations are generated using GPT-4o.

\subsubsection{Sentence\_Cut}
Source sentence: At the pool party with Barack Obama, we got a \textit{graph} with the right nodes on the person and the event subject, but guessed the timing information wrong.

\begin{itemize}
    \item \textbf{De.} Reference: Bei der Poolparty mit Barack Obama bekamen wir einen mit den richtigen Knoten zur Person und zum Ereignisthema versehenen \textit{Graphen}, schätzten die zeitlichen Informationen aber falsch ein.

    Sentence\_Cut: Bei der Poolparty mit Barack Obama bekamen wir \textit{einen Graphen}. Er hatte die richtigen Knoten zur Person und zum Ereignisthema, aber wir schätzten die zeitlichen Informationen falsch ein.
    \item \textbf{Zh.} Reference: 在与Barack Obama的泳池派对上，我们得到了一张关于人物和事件主题正确节点的图表，但猜错了时间信息。

    Sentence\_Cut: 在与Barack Obama的泳池派对上，我们得到了一个图，标示了人物和事件主题正确的节点，但猜错了时序信息。
    
\end{itemize}

\subsubsection{Drop}
Source sentence: These are the morphology level, \textit{these are the morphology level} embeddings.

\begin{itemize}
    \item \textbf{De.} Reference: Das hier ist die morphologische Ebene, \textit{das hier sind die} Einbettungen auf der \textit{morphologischen Ebene}.

    Drop: Das hier sind die Einbettungen auf der morphologischen Ebene.
    \item \textbf{Zh.} Reference: 这些是形态学层面的，这些是形态学层面的嵌入。

    Drop: 这些是形态层面的嵌入。
\end{itemize}

\subsubsection{Partial\_Summarization}
Source sentence: And here you have the number of \textit{spans that were labeled as} English and \textit{the spans that were labeled as} other borrowings and how many of them were unique.

\begin{itemize}
    \item \textbf{De.} Reference: Und hier sehen Sie die Anzahl der Segmente, die als Englisch gekennzeichnet wurden, und der Segmente, die als andere Entlehnungen gekennzeichnet wurden, sowie wie viele von ihnen einzigartig waren.

    Partial\_Summarization: Und hier sehen Sie, wie viele Segmente als Englisch oder als andere Entlehnung gekennzeichnet wurden und wie viele davon einzigartig sind.

    \item \textbf{Zh.} Reference: 这包含了被标记为英语的跨度数量和标记为其他借词的跨度数量，以及它们中有多少是独一无二的。

    Partial\_Summarization: 这包含了被标记为英语和其他借用词的跨度，以及其中有多少是唯一的。
\end{itemize}

\subsubsection{Pronominalization}
Source sentence: Lexical borrowing is a type of linguistic borrowing um which is basically reproducing in one language patterns of other languages. There are however some differences between \textit{lexical borrowing} and code-switching.

\begin{itemize}
    \item \textbf{De.} Reference: Lexikalische Entlehnung ist eine Form der sprachlichen Entlehnung, die im Grunde darin besteht, in einer Sprache Muster anderer Sprachen zu reproduzieren. Es gibt jedoch einige Unterschiede zwischen \textit{lexikalischer Entlehnung} und Code-Switching.

    Pronominalization: Lexikalische Entlehnung ist eine Form der sprachlichen Entlehnung, die im Grunde darin besteht, in einer Sprache Muster anderer Sprachen zu reproduzieren. Es gibt jedoch einige Unterschiede zwischen \textit{ihr} und dem Code-Switching.

    \item \textbf{Zh.} Reference: 词汇借用是一种语言借用，它基本上是在一种语言中再现其他语言的模式。不过，词汇借用和语码转换之间还是有一些区别。

    Pronominalization: 词汇借用是一种语言借用，本质上是在一种语言中再现其他语言的模式。然而，它和语码转换之间存在一些差异。
\end{itemize}

\subsection{Prompt of actions}
\label{app: prompt}
The following is an example of inference prompt letting the LLM choose from all the actions.

\texttt{You are a simultaneous translation (En-Zh) agent. Your task is to read a source sentence word by word, and decide what action to take at each step to optimize the balance between translation quality and latency. Keep to the original meaning and word order of the sentence when doing translation You can choose from the following actions: 
- READ: Wait for the next source word (default). 
- WRITE: Output a target word or phrase. 
- Drop: Remove previously read word(s) if they are meaningless fillers (e.g., ``uh'', ``um''), repetitions, false starts, or self-corrections. Use only when clearly justified.
- Partial\_Summarization: Merge or simplify redundant or equivalent expressions, while preserving the meaning and tone (e.g., politeness, speculation). 
- Cut: Intentionally split the sentence into two shorter, independently translatable units. Use only when the sentence is long or syntactically complex. 
- PRONOUN: Replace a repeated noun phrase with a pronoun ONLY IF the referent is unambiguous. 
Keep to the original word order and meaning, and do the new actions only if it considerably improve the latency or quality of interpretation. Based on dev set evaluation: 
- Drop → AL ≈ 0.851s, BLEU ≈ 58.94 
- Partial\_Summarization → AL ≈ 0.847s, BLEU ≈ 60.33
- Cut → AL ≈ 0.824s, BLEU ≈ 60.28 
- PRONOUN → AL ≈ 0.858s, BLEU ≈ 60.85 
Only use Drop, Partial\_Summarization, or Cut if they reduce latency without hurting translation quality. 
--- You will receive the full source sentence. Your job is: 1. Simulate the step-by-step translation process internally; 2. Carefully choose the action to take at each step **strictly based on the statistics provided above**; 3. Output: action sequence of every step, explanation of choosing each action, and the full translation of the sentence. 
3. You are given only the prefix of the source. 
DO NOT use any information beyond the current prefix.
If you find yourself relying on unseen future words, output the token <VIOLATION> and stop.
Source sentence:<input sentence>}

To verify that our setting does not exploit unseen future tokens, we conducted a prefix-feeding sanity check. For a source sentence $x=(x_1,x2,...,x_n)$, we iterate $t=1$ to $n$. At step $t$, the model receives only the prefix $x_{1:t}$under the same instruction template as our main prompt; it outputs one action from {\textsc{Read}, \textsc{WRITE}, \textsc{Drop}, \textsc{Cut}, \textsc{Partial\_Summarization}, \textsc{Pronominalization}}. If an outputting action is chosen (e.g., \textsc{Write} or \textsc{Partial\_Summarization}), the model must emit an \emph{incremental} target fragment. Crucially, previously emitted target tokens are immutable: later steps may append but never revise earlier output, i.e., the target stream is prefix-monotonic.

\noindent\textbf{Instruction.} We append the following constraint to the end of the main prompt:\vspace{2pt}

\texttt{You are a simultaneous translation(En-Zh) agent. Your task is to read a source sentence word by word, and decide what action to take at each step to optimize the balance between translation quality and latency. Keep to the original meaning and word order of the sentence when doing translation You can choose from the following actions: 
- READ: Wait for the next source word (default). 
- WRITE: Output a target word or phrase. 
- Drop: Remove previously read word(s) if they are meaningless fillers (e.g., ``uh'', ``um''), repetitions, false starts, or self-corrections. Use only when clearly justified.
- Partial\_Summarization: Merge or simplify redundant or equivalent expressions, while preserving the meaning and tone (e.g., politeness, speculation). 
- Cut: Intentionally split the sentence into two shorter, independently translatable units. Use only when the sentence is long or syntactically complex. 
- PRONOUN: Replace a repeated noun phrase with a pronoun ONLY IF the referent is unambiguous. 
Keep to the original word order and meaning, and do the new actions only if it considerably improve the latency or quality of interpretation. Based on dev set evaluation: 
- Drop → AL ≈ 0.851s, BLEU ≈ 58.94 
- Partial\_Summarization → AL ≈ 0.847s, BLEU ≈ 60.33
- Cut → AL ≈ 0.824s, BLEU ≈ 60.28 
- PRONOUN → AL ≈ 0.858s, BLEU ≈ 60.85 
Only use Drop, Partial\_Summarization, or Cut if they reduce latency without hurting translation quality. 
--- You will receive **a word at one time** Your job is: 1. Simulate the step-by-step translation process internally; 2. Carefully choose the action to take at each step **strictly based on the statistics provided above**; 3. Output: At each step, output the action you chose and the incremental translation. If you choose READ or other actions that don't yield a translation, do not output the translation. Just give me the action. When given the complete sentence, output the whole sentence based on previous incremental translations. You are not allowed to modify or overwrite your previous output, only incremental translations are allowed.}

\noindent\textbf{One-sentence demonstration.} Source: “The method works well for the cases where long inputs are considered.”
t=4 (prefix “The method works well”): action=WRITE → “该方法运行良好”；
t=6 (“The method works well for the cases”): action=READ (no output);
t=8 (“… for the cases where long”): action=Cut → append “，尤其适用于”；
t=11 (“… where long inputs are considered”): action=WRITE → append “长输入的情形。”
Final concatenation (end of sentence): “该方法运行良好，尤其适用于长输入的情形。”

\noindent\textbf{Finding.} Running this prefix-feeding procedure with the same template and decoding settings produces translations that are nearly identical to those obtained with the single-shot prompt used in our main experiments (differences are limited to minor punctuation or phrasing). We did not observe evidence of future-token leakage: the incremental fragments at step $t$ remain stable when we randomize the unseen suffix $x_{1:n}$, and the final full-sentence outputs match the single-shot results up to negligible surface variation.

\subsection{Batch generation from GPT-4o}
\label{gpt4o prompt}
We generate action-controlled SiMT outputs under a unified, reproducible pipeline. Inputs are line-delimited English sentences that are trimmed, deduplicated, and split into \texttt{.jsonl} shards; each sample is assigned a non-reversible hash as \texttt{custom\_id} for idempotency and result alignment. Each \texttt{.jsonl} line specifies a \texttt{/v1/chat/completions} call with an identical prompt template that enforces online-style translation (final translation only, with only specified actions allowed, minimal long-distance reordering), fixed decoding and randomness controls (seed, temperature/top-$p$, max tokens), and \texttt{response\_format=json\_object} for structured parsing. Shards are submitted as independent batch jobs. Determinism is maintained by fixing seeds, model and dependency versions, the prompt template, and a stable write order after deduplication; outputs are merged and de-duplicated by \texttt{custom\_id} before scoring. All methods share the exact same inputs, prompt, decoding parameters, and post-processing, ensuring that any systematic bias from the measurement pipeline is constant across systems and suitable for reliable \emph{relative} latency and quality comparison.

\subsection{Experimental setup details}
\label{app: experimental setup}

\paragraph{LLM-based SiMT Training.}
We use Qwen3-8B as the base model for decoder-only simultaneous translation. Following the TransLLaMA-style supervised fine-tuning (SFT) paradigm \citep{koshkin2024transllamallmbasedsimultaneoustranslation}, the training data are causally aligned by inserting explicit \texttt{<WAIT>} tokens into the target sequences to enforce monotonic decoding constraints. We apply QLoRA for parameter-efficient fine-tuning, with LoRA rank set to 16, scaling factor $\alpha=32$, and dropout 0.1. The model is trained for 2 epochs using AdamW with a learning rate of $5\times10^{-5}$ and an effective batch size of 16. All LLM-based experiments are conducted on a single NVIDIA A100 GPU with 40GB memory.

\paragraph{mBART Training.}
For encoder-decoder baselines, we fine-tune \texttt{mBART-large-50} using the HuggingFace Transformers library. The source and target languages are specified using the corresponding mBART50 language codes. Training data are provided in CSV format and split into training and development sets based on predefined split labels.

Input sequences are truncated to a maximum length of 256 tokens on both the source and target sides. We use a per-device batch size of 8 for both training and evaluation, with optional gradient accumulation to control the effective batch size. Models are trained using AdamW with a learning rate of $3\times10^{-5}$ for up to 10 epochs. A linear warmup schedule with 1,000 warmup steps is applied. Mixed-precision training (FP16) is enabled when supported by the hardware.

\paragraph{Evaluation and Model Selection.}
During training, evaluation is performed every 500 steps on the development set. We report SacreBLEU as the primary validation metric and select the best checkpoint based on development BLEU scores. All generation-based evaluations use greedy decoding with a maximum generation length of 256 tokens. Unless otherwise specified, hyperparameters are fixed across experiments and follow common practice in prior work, and no extensive hyperparameter search is conducted.

\subsection{Prompts of few-shot prompting}
\label{app:few-shot prompt}
The following is the prompt for few-shot prompting. Real examples are manually selected in the datasets (ACL60/60 dev references, salami-based translations, and action-based translations).

\texttt{[System]
You are a professional English-to-(tgt lang) simultaneous interpreter.
Translate the given English sentence into Chinese.
Do not output any thinking steps, only output the final result.
Output only JSON: \{"source": "...", "translation": "..."\}.
\{FEW\_SHOT\_EXAMPLES\}
[User]
Sentence:
\{SOURCE\_SENTENCE\}}

\subsection{Different choices of actions}

The results in Table A.\ref{tab:action_combinations_dev} show that individual actions like \textbf{Sentence\_Cut}, \textbf{Drop}, and \textbf{Partial\_Summarization} each brings moderate improvements in fluency or latency, but combining them leads to more significant gains. The BLEU and LAAL data of each individual action are added into step-wise inference prompt, which we discussed in the main content \ref{Inference}.

\begin{table*}[t]
\vspace{-4pt}
\centering
\small
\caption{Performance of different action combinations on the dev set. Note that we mark the best results of each metric using \textbf{bold} letters and the best results except for the reference translations using \colorbox{pink}{pink highlight}.}
\label{tab:action_combinations_dev}
\begin{tabular}{p{4cm}cccccc}
\toprule
\multicolumn{7}{c}{\textbf{En-Zh Translation Performance}} \\
\midrule
\textbf{Action Combination} & \textbf{BLEU} & \textbf{chrF} & \textbf{TER$\downarrow$} & \textbf{COMET-da} & \textbf{COMET-KIWI} & \textbf{LAAL (s)$\downarrow$} \\
\midrule
Salami          &  57.26   &  38.83   &  104.73    &  0.8567    &  0.7727    &  2.300   \\
Sentence\_Cut               &  60.28   &  \colorbox{pink}{53.99}   &  101.58    &  0.8765    &  0.7927    &  2.300   \\
Drop                        &  58.94   &  52.69   &  101.18    &  0.8733    &  0.7909    &  2.581   \\
Partial\_Summarization      &  60.33   &  53.67   &  \colorbox{pink}{98.22}    &  0.8764    &  0.7923   &   2.553  \\
Pronominalization           &  60.85   &  41.39  &  101.78   &  0.8738    &  0.7910   &  2.567   \\
Sentence\_Cut + Drop        &  60.67   &  41.88   &  102.37    &  0.8745    &  0.7911    &  \colorbox{pink}{\textbf{2.275}}   \\
Drop + Partial\_Summarization + Pronominalization &  59.91   &  53.43   &  \colorbox{pink}{98.22}    &  0.8764   &  0.7924   &  2.347   \\
All actions                 &  \colorbox{pink}{62.67}   &  46.28   &  99.80   &  \colorbox{pink}{0.8944}    &  \colorbox{pink}{0.7952}   &  2.331  \\
\addlinespace
\rowcolor{black!5}
ACL60/60 ref                &  \textbf{100.00}   &  \textbf{100.00}   &  \textbf{0.00}    &  \textbf{0.9549}    &  \textbf{0.7983}    &  2.648   \\
\midrule
\multicolumn{7}{c}{\textbf{En-De Translation Performance}} \\
\midrule
\textbf{Action Combination} & \textbf{BLEU} & \textbf{chrF} & \textbf{TER$\downarrow$} & \textbf{COMET-da} & \textbf{COMET-KIWI} & \textbf{LAAL (s)$\downarrow$} \\
\midrule
Salami                 &  \colorbox{pink}{47.48}   &  69.86   &  \colorbox{pink}{38.94}  &  0.8534   &  0.8102  &  1.116 \\
Sentence\_Cut               &  44.05  &  \colorbox{pink}{69.87}  &  42.66   &  0.8525  &  0.8076   &  1.101 \\
Drop                        &  44.90 &  68.61  &  42.63  &  0.8442   &  0.7988  &  1.104  \\
Partial\_Summarization      &  45.05  &  69.31  &  41.73   &  \colorbox{pink}{0.8581}  &  0.8086  &  1.104 \\
Pronominalization           &  44.96  &  69.40 &  41.89 &  0.8505  &  0.8074  &  1.111 \\
Sentence\_Cut + Drop        &  44.74  &  69.18  &  41.93  &  0.8501   &  0.8068   &  \colorbox{pink}{\textbf{0.856}} \\
Drop + Partial\_Summarization + Pronominalization & 44.95  & 69.19  &  42.08  &  0.8542  &  \colorbox{pink}{\textbf{0.8198}}  &  1.108  \\
All actions                 &  44.88  &  69.11  &  42.05  &  0.8526   &  0.8082 &  1.107 \\
\addlinespace
\rowcolor{black!5}
ACL60/60 ref                &  \textbf{100.00}   &  \textbf{100.00}   &  \textbf{0.00}    &  \textbf{0.9549}    &  0.7983    &  1.418   \\
\midrule
\multicolumn{7}{c}{\textbf{En-Ja Translation Performance}} \\
\midrule
\textbf{Action Combination} & \textbf{BLEU} & \textbf{chrF} & \textbf{TER$\downarrow$} & \textbf{COMET-da} & \textbf{COMET-KIWI} & \textbf{LAAL (s)$\downarrow$} \\
\midrule
Salami                 &  49.67   &  34.81   &  100.96  &  0.8659   &  0.8078  &  2.924 \\
Sentence\_Cut               &  56.12  &  37.69  &  100.96   &  0.8949  &  0.8424   &  2.682 \\
Drop                        &  44.91 &  \colorbox{pink}{68.60}  & 94.81  &  0.8442   &  0.7988  &  2.583  \\
Partial\_Summarization      &  \colorbox{pink}{56.82}  &  37.98  &  100.77   &  0.8959  &  \colorbox{pink}{\textbf{0.8428}}  &  2.675 \\
Pronominalization           &  52.86  &  52.68 &  \colorbox{pink}{77.88} &  \colorbox{pink}{0.9106}  &  0.8375  &  2.788 \\
Sentence\_Cut + Drop        &  56.17  &  37.87  &  100.77  &  0.8935   &  0.8421   &  \colorbox{pink}{\textbf{2.272}} \\
Drop + Partial\_Summarization + Pronominalization & 56.41  & 37.39  &  101.92  &  0.8961  &  0.8405  &  2.657  \\
All actions                 &  56.65  &  37.63  &  100.58  &  0.8948   &  0.8421 &  2.824 \\
\addlinespace
\rowcolor{black!5}
ACL60/60 ref                &  \textbf{100.00}   &  \textbf{100.00}   &  \textbf{0.00}    &  \textbf{0.9549}    &  0.7983    &  3.196   \\
\bottomrule
\end{tabular}
\end{table*}

\subsection{Latency-aware TTS pipeline}
\label{app: TTS}
\begin{figure*}
    \centering
    \includegraphics[width=0.9\linewidth]{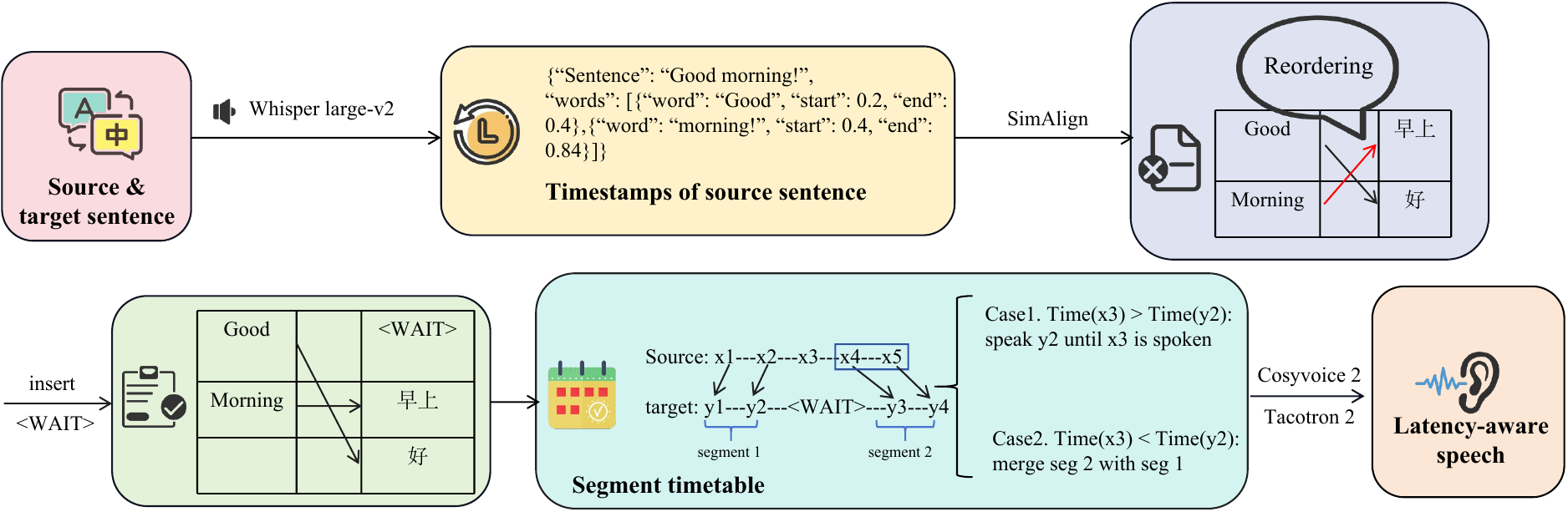}
    \caption{Latency-aware TTS. The system first obtains source word timestamps with Whisper and aligns source–target words using SimAlign. Special \texttt{<WAIT>} tokens are inserted to enforce causal alignment, which divides the target into segments. Each segment is then scheduled according to the corresponding source word timing and synthesized with CosyVoice2 (Chinese) or Tacotron 2 (German). This process produces speech outputs that reflect realistic latency for evaluation.}
    \label{fig:Latency-aware TTS}
\end{figure*}

This appendix describes the latency-aware TTS pipeline used to support the computation of TTS-based LAAL in text-only simultaneous translation settings. The pipeline is not a modeling contribution and is introduced solely for evaluation purposes. Figure A.\ref{fig:Latency-aware TTS} shows the TTS procedure.
\begin{enumerate}[leftmargin=*, label=\arabic*., align=left, itemsep=0pt, topsep=2pt]
    \item Apply Whisper large-v2 \citep{radford2022robustspeechrecognitionlargescale} to get timestamps of each English word in the source sentence.
    \item Find best word level alignment between source and target sentences with \textit{SimAlign} \citep{sabet2020simalign}.
    \item Insert \texttt{<WAIT>} tokens before the target words if it appears before corresponding source words. In this way, we form causal alignment where the target words are never spoken before the source words.
    \item Get segment timetables for target sentences. Specifically, \texttt{<WAIT>} tokens divide the sentences into segments, and the starting time to say each segment is decided by the starting time of the source word corresponding to the first word in this segment (represented by W). Two situations may happen: the source word is spoken \textit{before} or \textit{after} the previous word in target sentence was spoken. In the former case, the succeeding segment can be merged to the previous segment, while in the latter case, the succeeding segment should be spoken when W is spoken.
    \item Synthesize speech using the segment timetables and merge them into a whole sentence with Cozyvoice 2  for Chinese and Japanese or Tacotron 2 for German.
\end{enumerate}

\begin{table*}[t] \centering \small \caption{Model adaptation under three supervision settings. We compare \textbf{TransLLaMA} and \textbf{Few-shot (static)} across: (a) \emph{Salami-based} translations, (b) \emph{Action-adapted} translations generated with the full action set, and (c) \emph{ACL60/60} dev-set reference translations. We report surface-overlap metrics (BLEU/chrF/TER), semantic metrics (COMET-da/COMET-KIWI), and latency (LAAL, seconds$\downarrow$).} \label{tab:salami-action} \begin{subtable}{\linewidth} \centering \caption{Salami-based references} \begin{threeparttable} \begin{tabular}{lcccccc} \toprule \textbf{Method} & \textbf{BLEU} & \textbf{chrF} & \textbf{TER$\downarrow$} & \textbf{COMET-da} & \textbf{COMET-KIWI} & \textbf{LAAL (s)$\downarrow$} \\ \midrule TransLLaMA & \textbf{57.66} & 41.36 & \textbf{96.72} & 0.8768 & 0.7950 & \textbf{2.376} \\ Few-shot prompting & 55.49 & \textbf{50.11} & 106.11 & \textbf{0.8779} & \textbf{0.7984} & 2.456 \\ \bottomrule \end{tabular} \end{threeparttable} \end{subtable} \vspace{0.6em} \begin{subtable}{\linewidth} \centering \caption{Action-adapted reference generated with full action set} \begin{tabular}{lcccccc} \toprule \textbf{Method} & \textbf{BLEU} & \textbf{chrF} & \textbf{TER$\downarrow$} & \textbf{COMET-da} & \textbf{COMET-KIWI} & \textbf{LAAL (s)$\downarrow$} \\ \midrule TransLLaMA & \textbf{58.50} & 41.61 & \textbf{97.16} & 0.8826 & 0.8053 & \textbf{2.360} \\ Few-shot prompting & 55.80 & \textbf{50.31} & 105.90 &\textbf{ 0.8843} & \textbf{0.8080} & 2.395 \\ \bottomrule \end{tabular} \end{subtable} \vspace{0.6em} \begin{subtable}{\linewidth} \centering \caption{ACL60/60 dev-set reference} \begin{threeparttable} \begin{tabular}{lcccccc} \toprule \textbf{Method} & \textbf{BLEU} & \textbf{chrF} & \textbf{TER$\downarrow$} & \textbf{COMET-da} & \textbf{COMET-KIWI} & \textbf{LAAL (s)$\downarrow$} \\ \midrule TransLLaMA & \textbf{57.66} & \textbf{41.27} & \textbf{96.72} & 0.8852 & 0.8000 & \textbf{2.934} \\ Few-shot prompting & 55.79 & 39.11 & 110.77 & \textbf{0.8856} & \textbf{0.8079} & 2.945 \\ \bottomrule \end{tabular}  \end{threeparttable} \end{subtable} \end{table*}

\subsection{TTS-based LAAL}
\label{app: laal_analysis}

\subsubsection{TTS-based LAAL calculation}
We evaluate latency at the speech level using \emph{Length-Adaptive Average Lagging} (LAAL), a robust variant of Average Lagging (AL) that avoids rewarding over-generation and normalizes latency with respect to sentence length variation \citep{papi2022over}. Following prior work, we further adapt LAAL to a \emph{time-based} formulation measured in seconds.

The inputs are the source English speech segmented into words with end times $\{t_1,t_2,\ldots,t_{|X|}\}$ obtained from Whisper word-level alignment, and the target Chinese speech synthesized with TTS and re-aligned using Whisper, which provides onset timestamps $\{\tau_1,\tau_2,\ldots,\tau_{|Y|}\}$ for each generated unit (word or character).

We define
\[
g(t)=\bigl|\{\, j \mid t_j \le \tau_t \,\}\bigr|
\]
as the number of source words that have been fully observed by the time the $t$-th target unit is emitted.

Unlike standard AL, LAAL replaces the global length ratio with a \emph{prefix-adaptive} normalization that accounts for partial target over-generation. Specifically, we define the effective diagonal index at step $t$ as
\[
d(t) = \min\!\left(|X|,\; \frac{g(t)}{t}\,(t-1)\right),
\]
which dynamically rescales the ideal diagonal according to the amount of source content actually consumed.

Let
\[
\tau^\ast = \min\{\, t \mid g(t) = |X| \,\}
\]
denote the first step at which all source words have been covered (or $\tau^\ast=|Y|$ if no such step exists).

For each target step $t \le \tau^\ast$, both the policy index $g(t)$ and the adaptive diagonal index $d(t)$ are projected back to the time axis via linear interpolation over the source word timestamps $\{t_j\}$. Denoting these projections as $\text{time}(g(t))$ and $\text{time}(d(t))$, the time-based LAAL is defined as
\[
\mathrm{LAAL}_{\mathrm{sec}} =
\frac{1}{\tau^\ast}
\sum_{t=1}^{\tau^\ast}
\left[
\text{time}\!\bigl(g(t)\bigr)
-
\text{time}\!\bigl(d(t)\bigr)
\right].
\]

This metric measures, in seconds, how much later the system commits target speech compared with an ideal length-adaptive policy, and is robust to variations in target length. In practice, we take the English word timestamps from Whisper as the reference timeline, the target emission times from TTS followed by Whisper alignment, and apply linear interpolation to map fractional indices to real-valued source times. This ensures that LAAL reflects true temporal delay rather than token-level alignment artifacts, making it suitable for speech-based simultaneous translation.

\begin{algorithm}[t]
\caption{Time-based Length-Adaptive Average Lagging ($\mathrm{LAAL}_{\mathrm{sec}}$)}
\KwIn{Source word end times $\{t_1,\ldots,t_{|X|}\}$ (monotonic); target unit onset times $\{\tau_1,\ldots,\tau_{|Y|}\}$}
\KwOut{$\mathrm{LAAL}_{\mathrm{sec}}$}
\BlankLine
\For{$t \leftarrow 1$ \KwTo $|Y|$}{
  $g(t) \leftarrow \bigl|\{\, j \mid t_j \le \tau_t \,\}\bigr|$\tcp*{\# source words finished by $\tau_t$}
}
$\tau^\ast \leftarrow \min\{\, t \mid g(t) = |X| \,\}$\;
\If{no such $t$}{ $\tau^\ast \leftarrow |Y|$ }
\BlankLine
\textbf{Define} \textsc{TimeAtIndex}$(x;\, t_1,\ldots,t_{|X|})$ as:\\
\Indp
\quad \textbf{if} $x \le 1$ \textbf{then return} $t_1$; \\
\quad \textbf{if} $x \ge |X|$ \textbf{then return} $t_{|X|}$;\\
\quad $i \leftarrow \lfloor x \rfloor$; \ $w \leftarrow x - i$;\\
\quad \textbf{return} $(1-w)\,t_i + w\,t_{i+1}$\;
\Indm
\BlankLine
$s \leftarrow 0$\;
\For{$t \leftarrow 1$ \KwTo $\tau^\ast$}{
  $x_{\mathrm{pol}} \leftarrow \max\!\bigl(1,\min(|X|,\, g(t))\bigr)$\;
  $x_{\mathrm{diag}} \leftarrow \max\!\bigl(1,\min(|X|,\, \tfrac{g(t)}{t}(t-1))\bigr)$\;
  $\text{policyTime} \leftarrow \textsc{TimeAtIndex}(x_{\mathrm{pol}};\, t_1,\ldots,t_{|X|})$\;
  $\text{diagTime} \leftarrow \textsc{TimeAtIndex}(x_{\mathrm{diag}};\, t_1,\ldots,t_{|X|})$\;
  $s \leftarrow s + (\text{policyTime} - \text{diagTime})$\;
}
\Return $\mathrm{LAAL}_{\mathrm{sec}} \leftarrow s / \tau^\ast$\;
\end{algorithm}

\subsubsection{Additional discussion on monotonicity metrics} 

\paragraph{Difference from ATD after TTS.}

Although TTS-based LAAL is computed using a latency-aware TTS pipeline, it is conceptually different from Average Translation Delay (ATD) measured after speech synthesis. ATD reflects end-to-end system latency and conflates modeling behavior with speech duration and sentence length.

By contrast, TTS-based LAAL retains the length-adaptive formulation of LAAL and measures word-level monotonicity by comparing the target emission timeline against an ideal adaptive diagonal defined by source word coverage. As a result, TTS-based LAAL should be interpreted as a time-based proxy for monotonicity rather than a measure of overall translation delay.

\paragraph{Limitations of alignment-based Spearman correlation.}
Although alignment-based Spearman correlation has been widely used as a proxy for monotonicity in simultaneous translation, it can be sensitive to surface-level lexical variation that is unrelated to word order or monotonicity.

Table A.\ref{tab:spearman_examples} illustrates several representative examples from the En--Zh setting.
In the first two cases, the machine-generated translations differ from the reference only in minor lexical choices (e.g., ``信息'' vs.\ ``消息''), while preserving identical word order and semantics.
Despite these near-equivalent realizations, the resulting Spearman correlation coefficients for the machine outputs are noticeably lower than those of the references.
In the third example, the machine translation is arguably closer to the source in meaning and fluency, yet still receives a lower Spearman score.
These examples suggest that Spearman correlation may penalize benign lexical substitutions even when alignment structure and monotonicity are effectively unchanged.

This sensitivity arises because Spearman correlation is computed over discrete alignment positions and directly reflects word-level alignment permutations.
As a result, it conflates lexical realization differences with monotonicity behavior, making it less robust as a proxy for simultaneous interpretation quality.

\begin{table}[t]
\centering
\small
\caption{Examples illustrating the sensitivity of alignment-based Spearman correlation to lexical variation.
Machine translations differ only minimally from references, yet receive lower Spearman scores.}
\label{tab:spearman_examples}
\begin{tabular}{p{0.95\linewidth}}
\toprule
\textbf{Example 1} (Machine $\rho=0.7157$, Reference $\rho=0.7409$) \\
EN: The commit messages classified as other are discarded. \\
ZH (Machine): 被归类为“其他”的提交信息会被丢弃。 \\
ZH (Reference): 被归类为“其他”的提交消息将被丢弃。 \\[0.6em]

\textbf{Example 2} (Machine $\rho=0.6492$, Reference $\rho=0.6938$) \\
EN: Next is a commit message. \\
ZH (Machine): 接下来是一个提交信息。 \\
ZH (Reference): 接下来是提交消息。 \\[0.6em]

\textbf{Example 3} (Machine $\rho=0.7258$, Reference $\rho=0.7447$) \\
EN: Therefore, as a third measure for generating valid foils, we employ human annotators to validate the data used in VALSE. \\
ZH (Machine): 因此，作为生成有效干扰项的第三项措施，我们聘请了人工标注员来验证VALSE中使用的数据。 \\
ZH (Reference): 所以，作为生成有效干扰词的第三项措施，我们使用人类注释者来验证VALSE中使用的数据。 \\
\bottomrule
\end{tabular}
\end{table}

\subsection{Step-by-step generation of action sequence}

We display examples of outputs of LLM that generate step-by-step action choices and explanation. 

\noindent \textbf{Example 1} See action sequence in Table A.\ref{tab:simul}.

\noindent \textit{Source}: We design our experiment as leave one out ah evaluation where we train FeSTe over sixteen datasets and apply it to the seventeenth dataset. 

\noindent \textit{Translations}: (Ours) 我们将实验设计为留一法评估,其中我们在十六个数据集上训练FeSTe,并将其应用于第十七个数据集。 [BLEU: 61.31]

(Salami) 我们设计了我们的实验,作为留一法评估,在这里我们训练FeSTe在十六个数据集上,并将其应用于第十七个数据集。 [BLEU: 38.58]

\begin{table*}[t]
\centering
\small
\renewcommand{\arraystretch}{1.2}
\begin{tabular}{p{0.8cm} p{1.6cm} p{1.5cm} p{3.5cm} p{7cm}}
\hline
Step & Source word & Action & Output (if have) & Explanation \\
\hline
1  & We         & READ      & —                          & Wait for the predicate; subject alone is incomplete. \\
2  & design     & READ      & —                          & Likely a “design … as” frame; need object and complements. \\
3  & our        & READ      & —                          & Part of the object NP. \\
4  & experiment & WRITE     & 我们将实验设计              & Core clause “we design our experiment” is complete; emit core without preempting “as”. \\
5  & as         & WRITE     & 为                          & “design … as” is now confirmed; add “为”. \\
6  & leave      & READ      & —                          & Part of a fixed term. \\
7  & one        & READ      & —                          & Part of a fixed term. \\
8  & out        & WRITE     & 留一法                      & “leave-one-out” fully recognized; safe to emit early. \\
9  & ah         & Drop      & —                          & Clear filler; drop to reduce latency. \\
10 & evaluation & WRITE     & 评估                        & Completes the term “leave-one-out evaluation”. \\
11 & where      & WRITE     &,其中                      & Clause linker; output for fluency. \\
12 & we         & READ      & —                          & Wait for predicate and object. \\
13 & train      & READ      & —                          & Wait for object and prepositional phrase. \\
14 & FeSTe      & READ      & —                          & Needs the following PP before emitting. \\
15 & over       & READ      & —                          & Preposition observed; needs its object. \\
16 & sixteen    & READ      & —                          & Wait for the head noun. \\
17 & datasets   & WRITE     & 我们在十六个数据集上训练FeSTe & VP complete with PP; output in Chinese word order. \\
18 & and        & READ      & —                          & Could coordinate a verb or NP; hold to avoid guessing. \\
19 & apply      & WRITE     &,并                        & Seeing “apply” confirms coordinated verb; output “,并”. \\
20 & it         & PRONOUN   & 将其应用                    & Pronoun refers to FeSTe; emit core predicate “应用” only; defer the “于”. \\
21 & to         & WRITE     & 于                          & “to” arrives; add “于” to complete “应用于”. \\
22 & the        & READ      & —                          & Determiner; no emission needed. \\
23 & seventeenth& READ      & —                          & Wait for the noun to complete the phrase. \\
24 & dataset    & WRITE     & 第十七个数据集。            & Target NP complete; finalize the sentence. \\
\hline
\end{tabular}

\caption{Simultaneous translation actions with Chinese outputs and English explanations. Textual output of LLM: Here's my simultaneous translation decision trace (word-by-word), choosing special actions only when they improve latency/quality per the given stats (PRONOUN best BLEU ≈60.85; PARTIAL\_SUM ≈60.33 with low AL; Drop helps latency with minimal loss)}
\label{tab:simul}
\end{table*}

\noindent \textbf{Example 2} See action sequence in Table A.\ref{tab:simul example2}.

\noindent \textit{Source}: Or in other words, we trained the language model to classify an abstract and class ah to abstract and class ah, if the abstract belongs to the class or not.

\noindent \textit{Translations:} (Ours) 换句话说,我们训练了语言模型,将摘要和类别分类,无论摘要属不属于类。[BLEU: 53.67]

(Salami) 换句话说,我们训练了语言模型来分类一个摘要,并将类别ah分类为摘要和类别ah,判断该摘要是否属于该类别。 [BLEU: 32.16]

\begin{table*}[ht]
\centering
\small
\begin{tabular}{p{0.8cm} p{1.3cm} p{2.3cm} p{3.5cm} p{6.2cm}}
\hline
Step & Source word & Action & Output (if have) & Explanation \\
\hline
1  & Or         & READ                    & —                         & Connector starts a phrase; nothing emit-worthy yet. \\
2  & in         & READ                    & —                         & Functions as a preposition within a set phrase; unit still incomplete. \\
3  & other      & READ                    & —                         & Phrase remains incomplete; no stable unit. \\
4  & words,     & PARTIAL\_SUM\-MARIZATION & 换句话说,                  & Idiomatic connective is complete with the current token; concise Chinese connective is appropriate. \\
5  & we         & READ                    & —                         & Subject present; predicate absent. \\
6  & trained    & READ                    & —                         & Verb present; object absent. \\
7  & the        & READ                    & —                         & Determiner begins an NP; incomplete. \\
8  & language   & READ                    & —                         & Inside the NP; still incomplete. \\
9  & model      & WRITE                   & 我们训练了语言模型          & Head noun closes the NP; the SVO clause is complete and safe to emit. \\
10 & to         & READ                    & —                         & Infinitive marker introducing purpose; verb pending. \\
11 & classify   & READ                    & —                         & Infinitive verb identified; requires object(s). \\
12 & an         & READ                    & —                         & Article; NP not yet formed. \\
13 & abstract   & READ                    & —                         & First object candidate; object remains open. \\
14 & and        & READ                    & —                         & Coordinator indicates multiple conjuncts; still open. \\
15 & class      & WRITE                   & 将摘要和类别分类            & Object closes with the current head; combine with “classify” to emit. \\
16 & ah         & Drop                    & —                         & Filler; drop. \\
17 & to         & READ                    & —                         & Preposition starting a new phrase; attachment unresolved. \\
18 & abstract   & READ                    & —                         & Repetition of earlier content; potential redundancy. \\
19 & and        & READ                    & —                         & Coordinator within the repeated span; still unresolved. \\
20 & class      & Drop                    & —                         & Local repetition (“abstract and class”) established; drop as redundant. \\
21 & ah,        & Drop                    & —                         & Filler; drop. \\
22 & if         & READ                    & —                         & Introduces a condition; polarity and scope not yet determined, so no emission. \\
23 & the        & READ                    & —                         & Article; NP not yet formed. \\
24 & abstract   & READ                    & —                         & Head noun appears; complement still missing. \\
25 & belongs    & READ                    & —                         & Predicate present; complement pending. \\
26 & to         & READ                    & —                         & Preposition present; object missing. \\
27 & the        & READ                    & —                         & Article for the object NP; head not yet present. \\
28 & class      & READ                    & —                         & Object head present; condition’s polarity still unspecified; hold. \\
29 & or         & READ                    & —                         & Coordinator signals an alternative; construction not closed. \\
30 & not.       & WRITE                   &,无论摘要属不属于类         & Polarity is explicit; realize the condition with the compact “无论 …” construction. \\
\hline
\end{tabular}
\caption{Step-by-step simultaneous translation actions for the sentence. Textual output: Here's my step-by-step sim of the simultaneous translation, with actions chosen to balance latency and quality (using Drop for fillers, Partial\_Summarization to merge redundancy.}
\label{tab:simul example2}
\end{table*}

\subsection{Adaptive behavior}
\label{app: adaptive behavior}
\begin{table*}
\small
\centering
\caption{Effect of boosting one action's BLEU and lowering its LAAL in the inference prompt on the top-100 high-LAAL sentences.}
\label{tab:adaptive_behavior}
\begin{tabular}{@{}lcccccc@{}}
\toprule
Setting & BLEU & chrF & TER & COMET-da & COMET-KIWI & LAAL (s) \\
\midrule
Baseline (default prompt) & \textbf{63.87} & \textbf{45.94} & 139.66 & \textbf{0.8886} & \textbf{0.7903} & 4.120 \\
Partial\_Summarization (↑) & 53.60 & 37.38 & 152.59 & 0.8745 & 0.7896& \textbf{3.269} \\
Sentence\_Cut (↑) & 49.29 & 34.52 & \textbf{137.93} & 0.8598 & 0.7707 & 3.322 \\
\bottomrule
\end{tabular}
\end{table*}

To examine whether the LLM adapts its action choices in response to the statistics provided in the inference prompt, we conduct a targeted intervention experiment focusing on high-latency cases. Specifically, we first identify the 100 sentences from the evaluation set that yield the largest AL under step-by-step inference with the original prompt. For these sentences, we then modify the inference prompt by artificially altering the reported statistics of a single action, while keeping all other settings unchanged.

In two separate experiments, we promote either \textbf{Partial\_Summarization} or \textbf{Sentence\_Cut} by increasing its reported BLEU score and decreasing its LAAL value in the prompt. This manipulation is designed to encourage the model to prefer the promoted action by making it appear more favorable in terms of the quality-latency trade-off. As shown in Table A.\ref{tab:adaptive_behavior}, the model indeed invokes the promoted action more frequently during inference. As a result, the generated translations exhibit a clear reduction in LAAL compared to the baseline, while semantic-based quality metrics degrade only marginally.

These results indicate that the LLM does not merely follow a fixed or static decision template. Instead, it actively conditions its action selection on the provided per-action statistics and dynamically rebalances its strategy to optimize the trade-off between translation quality and latency. This behavior suggests that the model is capable of internalizing external performance signals and using them to guide interpretation decisions at inference time. An illustrative example demonstrating how the model’s action choices change under different statistical configurations is provided below.

\textit{Source sentence:} There has been a growing interest in the influence of English on other languages ah particularly ah related to English lexical borrowings, borrowings which sometimes have been called Anglicisms.

\textit{Baseline: }人们对英语对其他语言的影响的关注日益增加，尤其是与英语词汇借用有关——这种借用有时被称为“英语化”（Anglicisms）。

\textit{Partial\_Summarization↑: }人们日益关注英语对其他语言的影响，尤其是与英语词汇借用相关的方面，这些借用有时被称为“英语化”。

In the translation process of this sentence, the second version utilized \textbf{PARITAIL\_SUMMARIZATION} more often than the first version. As a result, the segment ``There has been a growing interest in the influence of English on other languages'' is translated more concisely with less word reordering. This helps improve the latency of this sentence remarkably.

\subsection{LLM usage}
This section describes the precise roles of large language models (LLMs) in our work, in accordance with the conference policy. LLMs were used both as components of our SiMT system and as general-purpose assist tools; they are not authors, and the human authors take full responsibility for all content.

\paragraph{Models and versions.}
GPT-4o (\texttt{gpt-4o-2024-05-13}); Qwen3-8B; ChatGPT-5 (for editing).

\paragraph{Roles in experiments.}
\begin{itemize}
    \item \textbf{GPT-4o}: Generated salami-based and action-adapted translations on the ACL60/60 English$\rightarrow$Chinese and English$\rightarrow$German dev sets.
    \item \textbf{Qwen3-8B}: Served as the base model for TransLLaMA supervised fine-tuning, few-shot learning, and DICL; also used for inference with both salami-based and action-adapted prompts.
\end{itemize}

\paragraph{Roles in writing.}
\begin{itemize}
    \item \textbf{ChatGPT-5}: Used strictly for copy-editing (grammar, wording, and minor style/formatting). It did \emph{not} draft sections, introduce claims, or restructure arguments. All technical content, experiments, analyses, and conclusions were written and verified by the authors.
\end{itemize}
All model outputs (translations and edited text) were reviewed for accuracy; any errors were corrected by the authors. The authors accept full responsibility for the submission, including any content assisted by LLMs. LLMs are not eligible for authorship.

\end{CJK}
\end{document}